# Using the power of memes: The Pepper Robot as a communicative facilitator for autistic children.

Linda Pigureddu
University of Turin, Italy, 332079@edu.unito.it
Cristina Gena
Dept. of Computer Science, University of Turin, Italy, cristina.gena@unito.it

This article describes the preliminary qualitative results of a therapeutic laboratory involving the Pepper robot, as a facilitator, to promote autonomy and functional acquisition in autistic children with low support needs (level 1 support). The lab, designed and led by a multidisciplinary team, involved 4 children, aged 11 to 13 years, and was organized in weekly meetings for the duration of four months. The following is the result of an in-depth qualitative evaluation of the interactions that took place between the children and the Pepper robot, with the aim of analyzing their effectiveness for the purpose of promoting the development of social and communication skills in the participants. The observations and analyses conducted during the interactions provided valuable insights into the dialogue and communication style employed and paved the way for possible strategies to make the robot more empathetic and engaging for autistic children.

## INTRODUCTION

Autism Spectrum Disorders take their name from the conception of the autistic condition as a heterogeneous spectrum, which can manifest in countless different ways. In the most recent edition of the DSM-5 (American Psychiatric Association, 2013 [1]), autism spectrum disorder is categorized as a neurodevelopmental disorder that involves features that impact the individual's social and communication skills. Although diagnostic manuals have moved beyond the former pathologizing view of autism, accepting that it is only a feature of mental functioning and not a mental illness[2], they recognize different levels of support for it (where level 1 is the lowest level of support, and level 3 is the highest level of support needed) that reflect the individual's autonomy in society and any co-occurrences with physical, psychiatric, and psychological impairments [3].

Recent studies on robot-assisted therapies for autism has produced very satisfactory results on the side of implementing social behaviors, spontaneous initialization of dialogue and management of stereotypies, often superior than traditional therapies [4]–[10]. This therapeutic success is credited to the special affinity that autistic children develop with the robotic partner, as it fulfills the needs for repetitiveness, stability and control of the environment typical in autistic minds, allowing the child to feel more comfortable in the laboratory setting and thus be more participatory during therapeutic activities [4]–[10]. Finally, it is underlined that for the children participating in the kind of projects it is clear that the robots cannot feel prejudiced toward them and are, as a result, more reassuring partners to them, which they can feel free to express themselves with without being limited by the unselfconscious dynamics of masking[1] [13]. This makes their use in the role of mediator within therapeutic settings particularly useful, as it encourages the dynamics of pairing allowing children to project their positive feelings toward the automaton upon the adults in charge associated with the robot, promoting social initiative and the success of plans aimed at the development of communication skills [14].

---

[1] Often also referred to as "camouflaging," in reference to ASD it is the ability of the autistic person to compensate for the deficits by hiding or limiting stereotypies (repetitive movements), internalizing discomforts related to sensory overloads, withholding meltdowns, i.e., explosive

This paper describes preliminary qualitative results from a therapeutic laboratory focused on the use of the robot Pepper to promote autonomy and functional acquisition in autistic children with low support needs (level 1 support). The analysis presented focuses primarily on the interactions and dialogues that occurred between the children and the robot, providing an in-depth view of the communicative dynamics within the laboratory setting. In addition, information is provided on the empirical results that emerged during the workshop phases, based on a qualitative analysis of the audiovisual material produced during the workshop. It is important to note that all workshop meetings were recorded, and evaluations were made based on these recordings.

## The project

**The laboratory**

Pepper was placed, in the role of mediator and dialogue partner, within the therapeutic laboratory[2] to promote autonomy and functional acquisition in autistic children with low support needs (level 1 support according to DSM-5 [1]). During the laboratory, conducted between the end of February 2021 and the beginning of June 2021, the exchanges, and interactions between autistic children in rehabilitation settings were explored, supported by the robot Pepper aiming to create an environment that would stimulate the development of social and communication skills, as well as strengthen the acquisition of strategies and autonomies related to daily activities, such as snack preparation [14]–[17].

Four autistic children with low support needs (level 1 support), aged 11 to 14 years, participated in the lab. During the lab sessions, the children had the opportunity to interact both directly with the robot and to participate in group activities, guided and coordinated by Pepper itself and the specialized operators, including educators, speech therapists, and psychologists [14]–[17].

The workshop sessions (16 in total), which were held weekly with a duration of two hours each, were recorded via a fixed camera, in addition to Pepper's 2D cameras (forehead camera and tablet camera), to allow post-hoc analysis to be conducted. During the meetings, two trainee students filled out evaluation forms provided by the psychotherapists, noting the progress of the sessions in a diary. These notes were later reworked to make reports, shared within the working group, reflecting the children's evolving behavior during their interactions with the robot [14]–[17]. To ensure the children's privacy, the "PepperForAutism" application was implemented in the robot during data collection, which consists of three main modules (Android app, web platform and backend) integrated into Jumple SRL's Pyramid platform. This system enabled the Pepper robot to uniquely identify the child with whom it was interacting, allowing data to be collected anonymously through association with a random ID known only to authorized individuals [14]–[17].

1.1. *The lab's sessions in details*

The sessions, lasting about two hours, were conducted by one or two specialized operators (educators, speech therapists, psychologists, etc.) assisted by two or three trainees. These were held in a room furnished with table and chairs and equipped with the tools necessary to carry out the activities, inside an apartment located in the center of Turin, made available by the Paideia Foundation also equipped with a bathroom, living room and kitchen. During the sessions, each child was given the opportunity to interact directly with Pepper

---

reactions and energetic manifestations resulting from uncomfortable situations that have become unbearable ([11]) and emulating neurotypical behaviors[12].

[2] Ethical approval for this study was obtained from the bioethical committee of the University of Turin, with approval number: 0664572.

and perform the activities planned for the group through the directions provided by the robot, with the help of the specialized operators and in a real context setting to leverage the benefit of a real-world evaluation [23]. Each session was phased and was conducted according to the following structure:

1. Welcome: children were made to sit in the apartment and welcomed by the therapists and the robot who, at the beginning of each session, provided directions on how to prepare to participate in the workshop.
2. First social time: dialogue session with the robot on a pre-established topic of interest (e.g., music, video games, etc.). The conversation is designed to be initialized by both the children and the robot. In the first case, "Getting Acquainted", the children would stand in front of the robot and, after waiting for it to catch their presence, proceed to ask a question of any sort. In the second case, identified as "Making Friends," Pepper would call the children one at a time to suggest a topic of conversation to each, according to a script updated weekly, in the expectation of engaging the other participants in a group conversation.
3. Development of practical skills (snack preparation and/or homework execution): Pepper played the role of activity coordinator, offering snack preparation instructions and explanatory videos displayed on the built-in tablet, in to guide the children in performing the task, with the therapists ready to intervene if needed. Starting from the 11th meeting, this phase was enriched by the inclusion of the activity of doing school homework. The children, who had been previously warned about the need to come equipped with their own school supplies, were asked to consult their journal to select a subject whose homework they wished to complete with the assistance of Pepper, who, after identifying a range of topics, proceeded by providing information on what materials were necessary or optional to proceed effectively with the task they chose. Again, the use of explanatory videos was provided.
4. Second social moment: post snack (or post homework) dialogue. The conversation is guided in the mode ("Moment of Knowledge" or "Making Friends") not explored in the First Social Moment.
5. Final Feedback: Before ending the session with greetings, Pepper asks the children to give their opinions about the meeting. In this phase, children were asked to stand in front of the robot and state their mood, choosing from the illustrated faces on the integrated tablet suitable for representing sadness, neutrality, and happiness. Because the workshop took place during a period characterized by social distancing due to the Covid-19 pandemic, children were required to wear a clear visor as an alternative to the surgical mask so that Pepper could take a picture, which was later used by the trainees to assess the consistency of the emotion indicated by the child with his or her facial expressions and the mood inferred by the robot [16].

1.2. *Welcome*

At the beginning of each meeting, Pepper welcomed the children and provided them with instructions for getting ready for and participating in the lab. These instructions included descriptive phrases and directives similar to those found in social stories, with the purpose of guiding and supporting the children during the activities. This phase is the one that produced the most noticeably positive results. Although Pepper offered to repeat the welcome directions each time, the children demonstrated that they remembered from one encounter to the next the rules of the lab and were able to correctly follow the instructions to take off their shoes and coats and to sanitize their hands before entering the lab, while still benefiting from the psychologist's help when necessary [17].

Starting therapeutic sessions in this way is both functional for the organizational and managerial purposes of the lab itself, and in line with the natural inclination of autistic minds to seek order and behavioral patterns that facilitate schematization of routines, predictability of events, and, in general, repetition [1]. This choice offers participants a sensitive approach toward the peculiarities of their minds, allowing them to feel immediately at ease, making it easier to introduce the activities planned for the day, and encourages children to be genuine and spontaneous, immediately comforting them by implicitly declaring that they are entering a safe environment, designed specifically to support and stimulate their minds, with no prejudice about their frailties.

### 1.3. *Practical skills training*

At this stage, the robot is assigned to coordinate the activity of snack preparation, offering snacks with recipes that were from time to time more and more complex, listing ingredients, encouraging children to get organized, providing guidance on the procedures both verbally and using pictures, animations and demonstration videos played on the built-in tablet, and giving children the opportunity to review the steps if necessary. During this activity, provision was made for Pepper to enter Waiting Mode, both to allow the children time to carry out the assignments and the opportunity for the group to spontaneously chat during the meal. On most occasions, the children took advantage of this time to share their suggestion for questions to ask Pepper during the conversational moments, and about activities and topics they would like to cover and proposal to improve the robot's dialogue structure to make it more friendly and pleasant to talk with. This was also because the time spent with Pepper was perceived as valuable, and the children were inclined to focus on it even when they were not communicating with it directly since they felt empowered with the responsibility and opportunity to contribute to the improvement of the robot.

The snack time produced good results and, in addition to being evidently stimulating for the young participants, was also one of their favorite moments; in fact, in most of the encounters they asked to start the lab session with this activity. The children showed appreciation for Pepper's help and, from the very beginning, good levels of autonomy, although the excitement associated with this time often affected the ability to maintain high concentration levels, as indicated by the scores given by the operatives [17].

Beginning in the eleventh meeting, the school homework activity was introduced, in which Pepper offered the children directions on what tools would be needed and tricks for doing the task they chose, always accompanying the verbal instructions with multimedia elements played on the integrated tablet when needed [17]. Regarding this phase, there was not sufficient data collected to assess its effectiveness, but it is still possible to consider the direct feedback from the children, who repeatedly and openly stated that they did not want to «waste» their time in the lab with ordinary activities, suggesting alternative skill-building activities (e.g., chemistry or robotics experiments), while only on one occasion one child did say that he appreciated the time devoted to doing homework, as it allowed him to associate the lab with the environment more familiar to him, of school.

### 1.4. *Conversations*

During the dialogue moments, the children were directly asked whether they wanted to "get acquainted" or "make friends" with the robot. In the former, they were given the opportunity to ask questions of their interest directly to Pepper, while in the latter, it was Pepper who asked them questions about their interests so as to encourage a collective conversation on the topic. Unfortunately, both modes did not produce satisfactory results, as they were too far removed from the human conversational style [20], consisting of simple repartee, without the possibility of in-depth or direct conversation development with the robot, due to the lack of an open conversational system that would allow the robot to use a communicative style more similar to natural language [17].

Giving children the opportunity to ask questions to the robot at a time when it lacked the ability to draw on ontologies to respond but had a dialogue system based on the manual and weekly update of its script, has produced various difficulties during the conversational moments. apart from not taking into account the difficulty of autistic people in initializing conversations [21], has produced several situations of misunderstanding, in which the robot, for technical difficulties (e.g. background noise, speech defects, etc.) or the absence of the input provided by the child within its script, it found itself unable to provide a coherent and satisfactory response [17].

After few interactions, the children began to ask themselves about Pepper with derisive attitude, deliberately asking nonsensical questions attempting to get funny answers or to evaluate how far they could go before sending cause a malfunction. An example of mocking interaction that has produced a lot of hilarity and has been revived during several meetings is related to the understanding by the children that pronounce at the beginning of the question the keywords «have you ever been in...» allowed access to the standard answer: «no, but I would like to visit the whole world», even though the question was not completed with a real location, prompting participants to ask if Pepper had ever visited the Moon, heaven or the land of the stupidity. As per the goal of stimulating the social interaction between the participants, the technical difficulties that emerged during this phase encouraged the children's curiosity about the robot, encouraging collective discussion about how Pepper works and brainstorming suggestions for improvements to make it a more enjoyable conversation partner. On the other hand, the analysis of these phases, despite not having produced satisfactory therapeutic results, is extremely interesting to conduct an evaluation of interactions aimed at improving the dialogue system, pointing out that, even before technical problems, communication with the robot is deficient because it is not fluent in the slang used by children, compromising its position as a mediator because it is unable to act as an interpreter between children and specialized adults working in the laboratory.

#### Memes

The qualitative analysis of the material produced during the project underlines that there is a linguistic barrier between Pepper and the children because Pepper's dialogue system was written by adults, while children talk through a vocabulary peculiar to their age. What's more relevant is that the participants integrate the use of memes into everyday conversations, which is quite surprising, even considering that's common practice between teens and pre-teens, if we consider the textbook definition of Autism Spectrum Disorders, since this requires the ability to master the underlying social context in the information presented just as a couple words-image matched through, usually ironic and sarcastic, rhetorical figures based on pop culture.

Although communicating through memes offers a few advantages that directly compensate for the difficulty autistic people face while entering social situations, such as:

- Regulate using echolalia in stressful and anxious situations, such as encountering people in public spaces, hidden in the repetition of a sound considered more socially acceptable, such as the meme's catchphrase.
- Transmission of disambiguous information: Memes, being a word-image pair, provide access to a concise and simplified non-verbal communication system that implies a broader meaning that cannot be misunderstood by those who know the sharing context. It provides a more accessible and comfortable way to express a concept than the risky natural communication that can cause misunderstandings.
- Sense of social belonging: the nature of a meme is that its information shared widely within online communities. This creates a sense of belonging to the social group of people who participate in certain platforms and applications. When a child enunciates a catchphrase, he assumes that his audience knows the meme from which it is taken and therefore understands what image it should be associated with, even though it is not shown (and vice versa). Receiving a response from the rest of the group constitutes an implicit expression of consent and confirmation of belonging to the same social group (those who know the meme), establishing mutual understanding and a friendly connection with others.
- Expression of emotions: memes are concentrated and concise formulas for the exposition of much broader, complex, and layered concepts, which, typically, autistic people find difficult to identify and express. This extreme simplification in a word-image pair allows children to identify their moods, more easily represent their emotional experiences, and communicate concepts that would be too complex for them to explain through traditional verbal language.

In the same way, these children usually communicate with each other using gestures referencing the images or videos presented in popular memes, expecting the group to catch the related tagline to understand what they're trying to express.

The usage of memes was so common during the laboratory that therapists suggested creating a list of "forbidden words", updated at the beginning of every encounter, containing all the words so overpronounced to be considered distracting from the children themselves. That ended up being a collection of meme catchphrases the participant admitted using so frequently to be an impairment to maintain a correct level of participation during the activities.

Updating the list at the beginning of every session prompted the children to be more self-conscious about their habits and what they found distracting; it also encouraged them to improve their vocabulary and communication skills to express themselves without using those memes. On the other hand, introducing this activity before starting with the program for the day helped them settle into the laboratory, slightly extending the familiarization process begun in the welcome phase. Taking advantage of echolalia, as the children's relief associated with the repetition of words and sounds that are reassuring to them, is comforting for them and makes them aware they are entering a safe space, where they will receive help to improve their skills in a way that understands the peculiarity of their minds and free of any prejudice.

According to this observation, in future applications, Pepper should be able to understand the meaning behind gestures, especially basic ones, such as hand raising, to catch non-verbal inputs, and when the children are referencing a meme and how to implement those into the dialog, to fully understand what is being communicated by the children and increase the perception of trust, credibility, and belonging to the same social group, helping the children to see Pepper as their peer and effectively use it in the role of mediator.

To implement this kind of information, it can be useful to involve a group of teenagers and pre-teens, or a content creator who specializes in making content for this age group, in the process of making the robot persona and creating the dialogues.

Another way to do this can be by leveraging machine learning strategies already planned to improve the quality of conversations, allowing the robot to learn directly from the children participating in the project, who demonstrated to be interested in helping with improving the robot and explaining their slang, recording the implicit meaning of a pun or a gesture for future reference and usage during the interactions.

The use of machine learning also opens up the opportunity to dip into the websites most frequently used by this age group, such as social networks and video game forums. This solution can be helpful to keep Pepper's vocabulary updated but is certainly more challenging due to the amount of data collected by this platform and the difficulty in designing a strategy to make the robot understand the subtext of the meme and to keep up with the evolution of the trends, even considering the possibility of relying on online databases that are updated frequently and that aim precisely to explain the meaning of this type of content, such as knowyourmeme.com.

Finally, it is important to consider that the robot will interact with children, augmenting the risk of mistakenly implementing inappropriate content into the robot, considering that the majority of the information stored online is not child friendly.

## Conclusion

The analysis of the interactions and dialogues that took place during the therapeutic laboratory for autonomies endorsed with the inclusion of the Pepper robot provided an interesting perspective on the communication dynamics of children in the lab.

The results obtained showed the effectiveness of the robot in promoting autonomy and functional acquisition, demonstrating that the therapeutic method based on assistive robotics is a valuable resource to support the rehabilitation needs of communication and social skills of autistic children. The analysis highlights new possibilities for their engagement and active participation as a co-designer, as we already experienced in the past with neuro-typical children [19], representing an innovative opportunity to promote the development and wellness of autistic people and opening to new perspectives for therapeutic intervention more aware of the needs of autistic minds.

In conclusion, this paper highlights the need to provide the robot with the ability of adapt to the children's peculiarity and features and sharing her/his vocabulary, especially by recognizing and using memes during interactions, to inspire greater trust in children and allowing the use of common slang, already in use with classmates, allowing them to consider Pepper a peer and effectively insert it in the role of mediator. In addition, it would allow children to use a simplified type of communication during meetings to compensate for the typical communication deficits associated with autism spectrum disorders.

In the future we will work in this directions (co-design, user's adaptations, using and recognizing memes during children-robot communication) having the robot able to communicate and express adapting to user tastes and preferences and and try again to field a real-world evaluation that takes into account the effectiveness of different levels of user adaptation [23].